\documentclass{article}

\usepackage[preprint]{neurips_2022}

\usepackage[utf8]{inputenc} 
\usepackage[T1]{fontenc}    
\usepackage{hyperref}       
\usepackage{url}            
\usepackage{booktabs}       
\usepackage{amsfonts}       
\usepackage{nicefrac}       
\usepackage{microtype}      
\usepackage{graphicx}
\usepackage[usenames,dvipsnames]{color}
\usepackage[title]{appendix}
\usepackage{natbib}

\title{Centralized control for multi-agent RL in a complex Real-Time-Strategy game}

\author{%
  Roger Creus Castanyer \\
  Mila Québec\\
  Université de Montréal\\
  \texttt{roger.creus-castanyer@mila.quebec} \\
}

\begin{document}

\maketitle

\begin{abstract}
Multi-agent Reinforcement learning (MARL) studies the behaviour of multiple learning agents that coexist in a shared environment. MARL is more challenging than single-agent RL because it involves more complex learning dynamics: the observations and rewards of each agent are functions of all other agents. In the context of MARL, Real-Time-Strategy (RTS) games represent very challenging environments where multiple players interact simultaneously and control many units of different natures all at once. In fact, RTS games are so challenging for the current RL methods, that just being able to tackle them with RL is interesting. 
This project provides the end-to-end experience of applying RL in the \textit{Lux AI v2} Kaggle competition, where competitors design agents to control variable-sized fleets of units and tackle a multi-variable optimization, resource gathering, and allocation problem in a 1v1 scenario against other competitors. We use a centralized approach for training the RL agents, and report multiple design decisions along the process. We provide the source code of the project: \url{https://github.com/roger-creus/centralized-control-lux} 
\end{abstract}

\section{Introduction}

Reinforcement learning (RL) has achieved great success in many complex domains, such as video games and robotics \citep{mnih2013playing}.
However, most of the work in RL assumes single-agent settings, where the agent interacts with a static or predetermined environment. In recent years, there has been growing interest in multi-agent RL (MARL), where multiple agents interact with a shared environment and learn to coordinate or compete. Many of the most recent and impactful RL successes use MARL. For example, OpenAI Five \citep{berner2019dota}, a team of five AI agents, defeated a team of professional human players in the complex video game Dota 2. Similarly, AlphaStar \citep{vinyals2019grandmaster}, a system developed by DeepMind, defeated professional human players in the StarCraft II game. AlphaGo \citep{silver2016mastering}, another DeepMind system, defeated the world champion human player in the board game Go. Recently, Meta AI published Cicero \cite{meta2022human}, an AI agent that outperforms humans in the game of Diplomacy. These successes demonstrate the potential of MARL and the importance of developing effective strategies for coordination and cooperation among agents in complex environments. 

\begin{figure}[h!]
\includegraphics[scale=0.2]{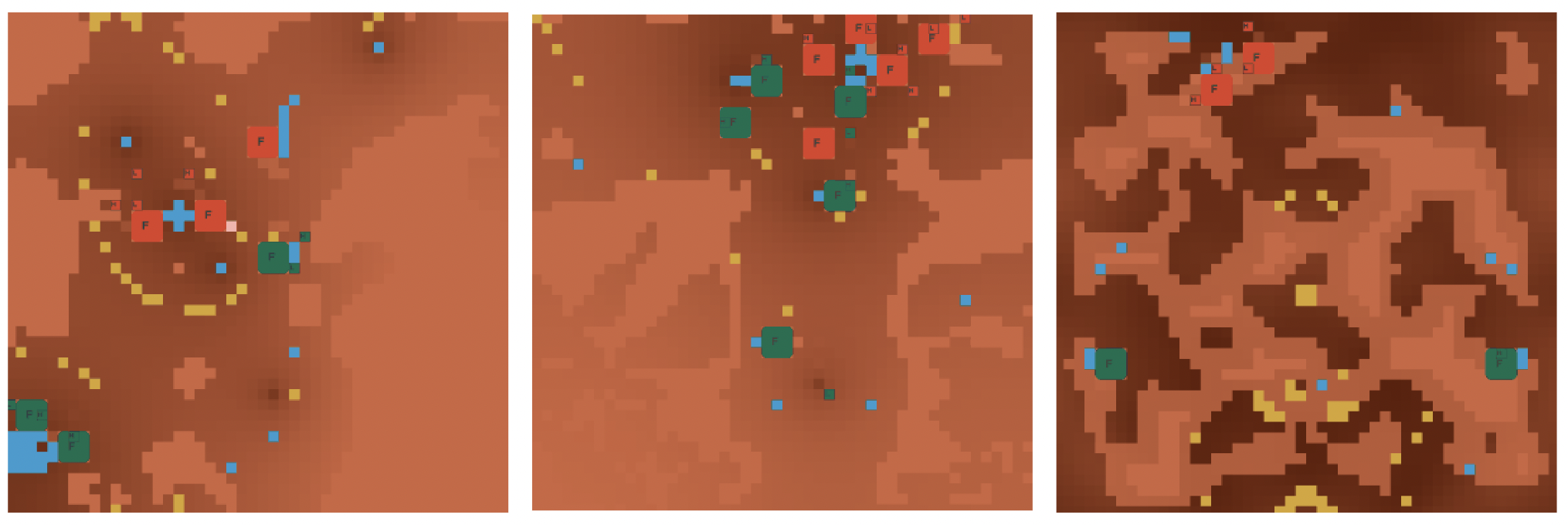}
\centering
\caption{Three different 48x48 maps of the \textit{Lux AI v2} environment. {\color{red} Player1} and {\color{ForestGreen} Player2} compete for resources in pseudo-randomly generated maps. Yellow cells indicate {\color{Dandelion} \textit{ore}}, blue cells indicate {\color{Aquamarine} \textit{ice}}, brown cells indicate {\color{Maroon} \textit{ruble}}, 3x3 squares with the letter F are \textit{factories}, and little squares with the letters L and H are \textit{light} and \textit{heavy} robots respectively.}
\label{lux}
\end{figure}

Due to the computational requirements for training such complex systems, the majority of research institutions can't contribute to MARL with significant empirical evidence. However, there have been many recent advances in the design of affordable but complex environments which pose similar challenges to StarCraft and Dota II \citep{huang2021gym, chen2023emergent, suarez2019neural}. Concretely in terms of coordination between units, size of the joint action space, non-stationary dynamics and non-Markovian rewards, and hence enable the research community in MARL to empirically study new techniques and algorithms. Several competitions have been hosted recently in such affordable environments because it remains challenging to tackle these with RL due to the implicit complexity of MARL and RTS games. In this work, we participate in the \textit{Lux AI v-2} Kaggle competition \footnote{\url{https://github.com/Lux-AI-Challenge/Lux-Design-S2}} and provide the end-to-end experience from defining the environment observation and action spaces to training RL agents via self-play. The \textit{Lux} environment can be seen in Figure \ref{lux}. Section \ref{lux-env} and the environment specifications\footnote{\url{https://www.lux-ai.org/specs-s2}} provide more details on Lux.

The Lux environment poses several key challenges: {\it{(i)}} Lux provides entity observations which need to be pre-processed before feedxing them to an RL agent (e.g. see a sample observation \href{https://github.com/Lux-AI-Challenge/Lux-Design-S2/blob/main/dev/sample_obs.json}{here}); {\it{(ii)}} at each turn, each player needs to issue primitive actions for variable-sized fleets of units; {\it{(iii)}} the units have different action spaces and specifications; {\it{(iv)}} resources are sparse and critical for survival; {\it{(v)}} there is a high variability of maps and terrains, the number of factories is also different at each game; and finally {\it{(vi)}} it's a non-stationary environment because it involves another agent. Arguably, these engineering, algorithmic and theoretical challenges conform to a very complicated environment for RL. In this work, we tackle each one of them and report our empirical experience.

\section{Background}
In the context of MARL, the Lux environment can be modelled as a Multi-Agent Markov Decision Process (MMDP). An MMDP \citep{boutilier1996planning, van2021multi} is a tuple $(\mathcal{N}, \mathcal{S}, \mathcal{A}, \mathcal{T}, \mathcal{R})$ where $\mathcal{N}$ is a set of \textit{m} agents, $\mathcal{S}$ is the state space, $\mathcal{A} = \mathcal{A}_1 \times \dots \times \mathcal{A}_m$ is the joint action space of the MMDP, $\mathcal{T} : \mathcal{S} \times \mathcal{A}_1 \times \dots \times \mathcal{A}_m \times \mathcal{S} \rightarrow [0, 1]$ is the transition function,
and $\mathcal{R} : \mathcal{S} \times \mathcal{A}_1 \times \dots \times \mathcal{A}_m \rightarrow \mathcal{R}$ is the immediate reward function. This \textit{decentralized} setup models the coordination between the \textit{m} agents in the same team. $\mathcal{S}$ is the global state space, and the transition function models the probability of the future global states under the m-dimensional joint action distribution. Similarly, this framework defines a global reward distribution as a function of a single global state and the joint \textit{m} actions, meaning that the \textit{m} agents collaborate to maximize a single reward function. Moreover, in a 1vs1 environment like Lux, the opponent's units' actions are part of the state space $\mathcal{S}$, and if the opponent is a learning agent or evolves with time, then the transition distribution $\mathcal{T}$ is non-stationary. 

Credit assignment in decentralized MARL is challenging \citep{chang2003all}. For this reason, several works explore value-decomposition methods \citep{sunehag2017value} to learn simpler agent-specific value functions or critics \citep{iqbal2019actor}. Other works leverage hand-crafted agent-specific reward distributions to independently train each agent \citep{ye2020towards}. However, such works assume a fixed number of agents and aren't feasible in massive environments like Lux where units are constantly destroyed and built. Finally, several works define a \textit{centralized} set-up and use a pixel-to-pixel architecture that leverages multi-agent control in grid-like environments \citep{huang2021gym, chen2023emergent, han2019grid}. In this context, the MMDP turns into a standard single-agent problem. However, the size of the action space, the cooperation between units, and the non-stationarity of the transition function remain challenging. In this work, we also use a centralized approach for training our RL agents, which we optimize with Proximal Policy Optimization (PPO) \citep{schulman2017proximal}. PPO is an actor-critic policy-gradient-based algorithm, which compared to vanilla REINFORCE or A2C \citep{grondman2012survey}, defines a surrogate objective which allows re-using the collected on-policy data for multiple corrected training updates. In Section \ref{meth}, we describe the model architecture in detail and the PPO training process.

\section{Methodology} \label{meth}
This section defines the components that enable training an RL agent in the Lux environment. The main requirements are defining the state and action spaces, a dense reward distribution, a functional model architecture, and a training pipeline. While these requirements are present in all RL applications, their complexity scales as the environment complexity grows, making RL challenging in RTS games. The following describes our design decisions, which motivation is to turn Lux into a fully-observed MDP that allows training a single, and thus centralized, RL agent.


\textbf{Observation space}. We define the observations by extracting several feature maps containing independent information about the game. The feature maps are stacked together as objects of size $(H \times W \times C)$ where $H=48$, $W=48$ is the fixed map size and in our case $C=17$. These provide information about the positions of the ally and enemy units, factories and resources on the map. Table \ref{tab:obs} contains a detailed description of the observation space.

\textbf{Action space}. We define two action spaces, one for the factories (which can create robots and grow lichen) and one for the robots (which can move, dig and transfer resources). The factory's action space is multi-discrete with 2 components: the first one indicates the position of the factory on the map and the second allows 4 actions. The unit's action space is multi-discrete with 6 components: the first one indicates the unit's position on the map and the other 5 define the joint action (e.g. transfer - ice - south). Section \ref{lux-env} contains detailed descriptions of the action space. Furthermore, invalid action masks \citep{huang2020closer} are implemented to prevent the agent from learning from output actions that are not going to be executed in the game - the model outputs an action at each cell of the 48x48 map, but the agent shouldn't learn from actions that will not be issued to an actual unit. The action masks are used to effectively set the logits of illegal actions to  large negative values so that their log probabilities are 0 and so are their gradients.

\textbf{Reward distribution}. We use reward shaping to define a hand-crafted reward distribution that encourages players to generate water. While players win by growing as much lichen in 1000 game steps, resource requirements are too critical in Lux to have RL agents learn to generate water only from game-winning rewards. To generate water, which is needed for survival and growing lichen, units have to navigate efficiently, dig ice, transfer it to factories and have it refined. Arguably, the latter is a too complex sequence of actions to be discovered from sparse or delayed rewards. A more detailed discussion of the reward distribution can be found in Section \ref{app-reward}.

\textbf{Self-play}. Prioritized Fictitious Self-play (PFSP) \citep{vinyals2019grandmaster} is implemented to provide a stable training pipeline for the RL agents. Compared to population-based approaches \citep{jaderberg2017population}, in PFSP a single learner agent plays against a pool of previous checkpoints of itself. The pool is implemented as a queue of fixed size, so after every $N$ training updates a new checkpoint is stored and the oldest one is removed. During training, the most recent checkpoint is sampled with a 50\% chance and any other one randomly otherwise. Section \ref{app-selfplay} discusses self-play in more detail.

\begin{figure}[h!]
\includegraphics[scale=0.3]{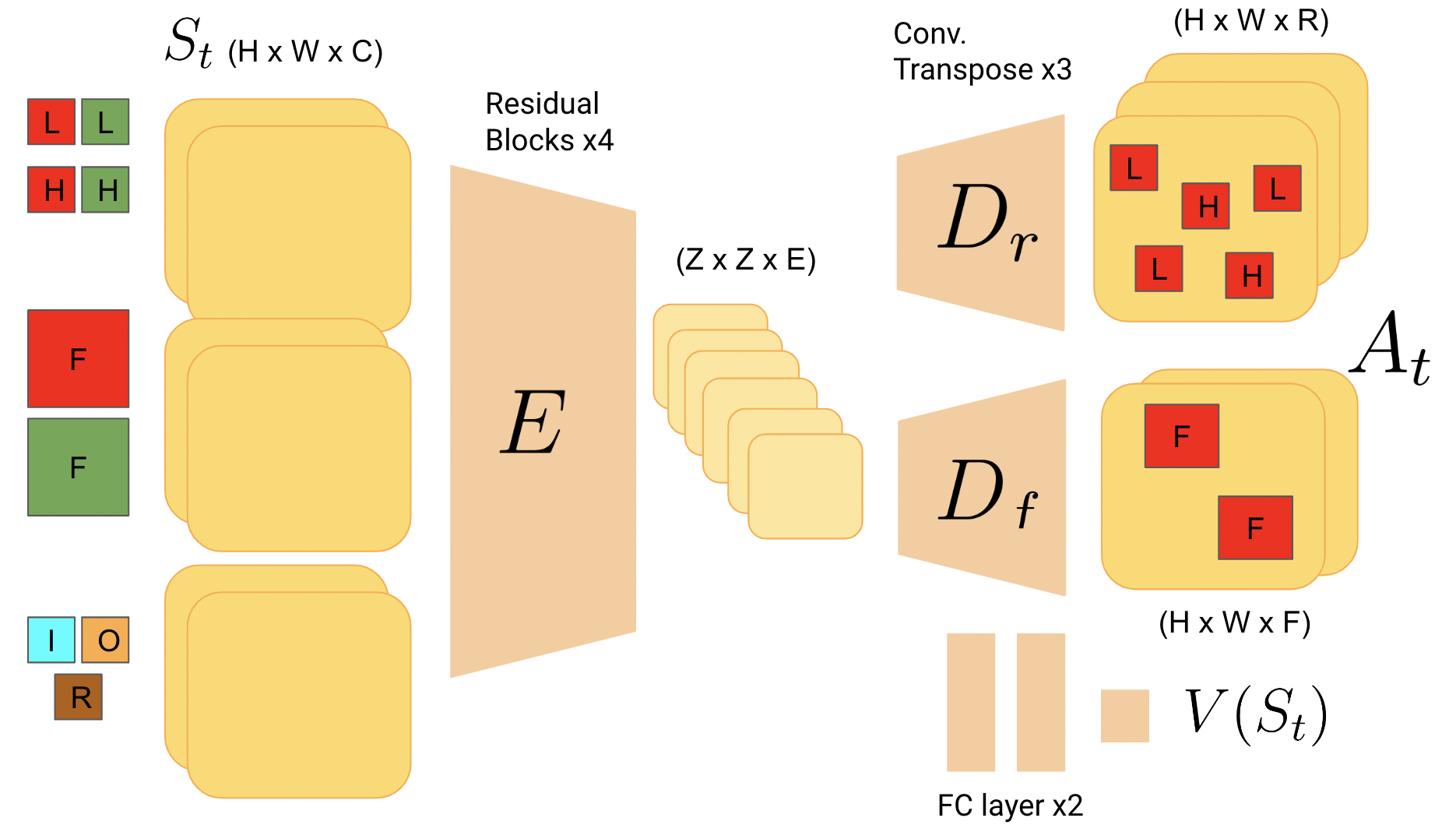}
\centering
\caption{Our pixel-to-pixel architecture. We feed the model with a stack of C feature maps containing information about the units, factories and resources. We use a Residual Network \citep{he2016deep} with squeeze-and-excitation layers \citep{hu2018squeeze} to encode the observations into low-dimensional representations. We feed the representations to three different heads: the critic which outputs a single scalar value representing the state value; the \textit{robot actor} which outputs feature maps with as many dimensions as robot action components; and the \textit{factories actor} which outputs as many actions as factory action components.}
\label{arch}
\end{figure}

\textbf{Architecture}. An actor-critic pixel-to-pixel architecture \citep{han2019grid} is used, which is similar to \textit{Gridnet} \citep{chen2023emergent}, and trained with PPO. The model architecture is shown in Figure \ref{arch}. During training, the model computes the joint action probabilities of all valid robot and factory actions by summing their logits. The latter corresponds to multiplications in the space of probabilities which assumes independence between factory and robot actions. The critic is used to compute the returns and advantages, and the PPO updates consist of making the actor's predicted actions with higher estimated advantages more likely.

\section{Experiments}

In general, RL needs lots of data to work,
and MARL is usually orders of magnitude more sample-inefficient \citep{garnelo2021pick, berner2019dota, vinyals2019grandmaster}. Furthermore, the Lux environment is slow, achieving 60 environment steps per second running on a single A6000 GPU. We perform a hyperparameter sweep\footnote{\url{https://api.wandb.ai/links/rogercreus/49pdlt7a}} running many jobs in parallel, each training the agent for 10M environment steps for a total of 1,296 hours of computing. After that, we take the best-performing configuration and run it using 8 RTX8000 GPUs in parallel for 220M environment steps during 124 hours. The report of the. More details can be found in Section \ref{app-performance}.

\begin{figure}[h!]
\includegraphics[width=\linewidth]{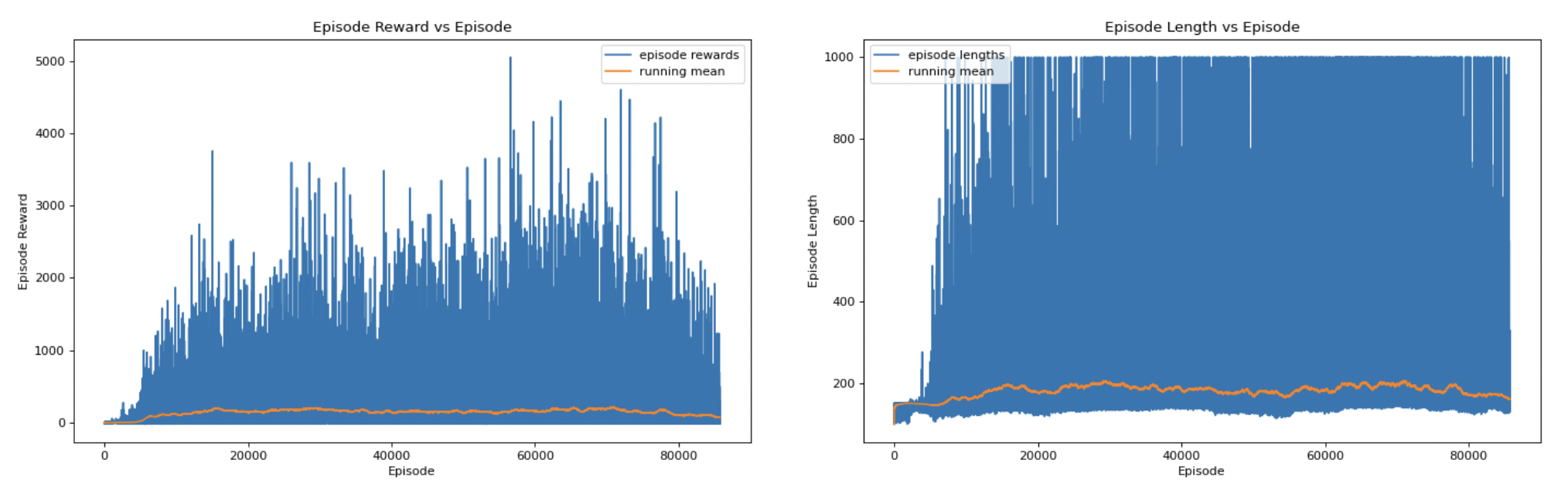}
\centering
\caption{The figure shows the episode reward (left) and episode length (right) during training. The agent is trained for 220M environment interactions. Although the agent learns to gather resources and survive for the 1,000 steps of the game in many episodes, there is a high degree of variance. Since each episode has a significantly different map and a variable number of factories, it is hard for the agent to generalize.}
\label{performance}
\end{figure}

\section{Conclusions and Future work}
This work defines a centralized training framework for MARL and shows how to train a pixel-to-pixel actor-critic architecture with PPO in a highly challenging environment for RL. We have uncovered several challenges for applying RL in RTS games, justified our design decisions and discussed possible improvements. We have also provided an open-source codebase which can be used as a starting point for similar future competitions \footnote{\url{https://github.com/roger-creus/centralized-control-lux}} and shared the logs for all the experiments\footnote{\url{https://api.wandb.ai/links/rogercreus/49pdlt7a}}.

Several lines of work can improve the presented approach, some of which are: {\it{(i)}} better definitions of the observation and action spaces; {\it{(ii)}} using other architectures whose inductive biases are more appropriate for grid-like RTS games (e.g. avoiding the downscaling bottlenecks \citep{chen2023emergent}, U-Net \citep{ronneberger2015u}, LSTM \citep{hochreiter1997long}); {\it{(iii)}} implementing population-based-training to shape the learning dynamics in favour of the learner agent \citep{garnelo2021pick}; and {\it{(iv)}} scaling up the architecture and enabling highly distributed training.

\clearpage

\bibliographystyle{plainnat}
\bibliography{references.bib}

\begin{appendices}
\section{The Lux-v2 competition} \label{lux-env}

In Lux, both AI agents and scripted bots compete for resources in 1vs1 games. The game has 3 phases: {\it{(i)}} the \textit{starting phase}, in which both players bid for who will place a factory first; {\it{(ii)}} the \textit{placement phase}, in which starting with the player who won the bid, each player iteratively places a factory at its chosen locations; and {\it{(iii)}} the \textit{normal phase}, in which both players issue actions to all the factories and robots to gather resources and grow lichen. In this work, we have heuristically set the \textit{bidding phase} to 0 for both players and the \textit{placement phase} to a strategy that places the factories taking into account the locations of ice, ore, rubble and enemy factories. In this way, in our environment wrapper, each call to \textit{env.reset()} already simulates the bidding and placement phases and returns the first state of the normal phase.

The games last for a maximum of 1,000 turns in which the two players submit actions simultaneously. If a player loses all their factories, the game ends. Otherwise, the player who has more lichen on the map wins. 

Factories can either build light or heavy robots (consuming metal and power) or grow lichen (consuming water). Robots can move in 4 directions (consuming power), transfer resources (ice, ore, water, metal and power) in 4 directions, dig resources (ice, ore, rubble), and pick up resources from factories or self-destruct. Furthermore, robots act according to action queues, which allow scripted agents to submit entire action plans in a single step. Updating the action queue of a robot has an additional cost of power, encouraging efficient planning of sequences of actions. In this work, we assume single actions, paying the additional cost in power.

\clearpage 

\section{Observation Space} \label{app-obs}

Lux originally provides entity observations, which are dictionaries (i.e. JSON files) containing all the information of the current game state. For this reason, we parse this information into a format that can be fed into a deep RL model and describe it in the following:

\begin{table}[h!]
\centering
\caption{Description of the 17 feature maps that conform the observations fed to the agent. Negative values always indicate properties related to the opponent (e.g. a -200 in unit\_ice means an enemy unit has 200 ice in it's cargo. Each of the feature maps is the same size of the map 48x48.}
\label{tab:obs}
    \begin{tabular}{|l|l|c|c|}
        \hline
        \textbf{Name} & \textbf{Description} & \textbf{Range} & \textbf{Normalization} \\
        \hline
        is\_day & Whether the current turn is day or night & [0, 1] & - \\
        rubble & Amount of rubble in each cell & [0, 100] & 100 \\
        ore & Whether there is ore in each cell & [0, 1] & - \\
        ice & Whether there is ice in each cell & [0, 1] & - \\
        lichen & Amount of lichen in each cell & [0, 100] & 100 \\
        is\_resource & Whether there is ore or ice in each cell & [0, 1] & - \\
        light\_units & Whether there is a light robot in each cell & [-1, 0, 1] & - \\
        heavy\_units & Whether there is a heavy robot in each cell & [-1, 0, 1] & - \\
        unit\_ice & Amount of ice in the robot's cargo & [-3000, 3000] & 3,000 \\
        unit\_ore & Amount of ore in the robot's cargo & [-3000, 3000] & 3,000 \\
        unit\_power & Amount of power in the robot's battery & [-1000, 1000] & 1,000 \\
        unit\_on\_factory & Whether each robot is on top of a factory & [-1, 0, 1] & - \\
        factories & Whether there is a factory in each tile & [-1, 0, 1] & - \\
        factory\_ice & Amount of ice in each factory & [-$\infty$, $\infty$] & 150 \\
        factory\_ore & Amount of ore in each factory & [-$\infty$, $\infty$] & 150 \\
        factory\_water & Amount of water in each factory & [-$\infty$, $\infty$] & 150 \\
        factory\_metal & Amount of metal in each factory & [-$\infty$, $\infty$] & 150 \\
        \hline
    \end{tabular}
\end{table}

\section{Action Space}

We define a joint action space for both robots and factories as a dictionary (\textit{gym.spaces.Dict()}) in which both \textit{action\_space["factories"]} and \textit{action\_space["robots"]} are MultiDiscrete objects. A MultiDiscrete space is the multi-dimensional extension of the simple Discrete space \footnote{\url{https://www.gymlibrary.dev/api/spaces/\#multidiscrete}}. In a MultiDiscrete action space, each component (i.e. dimension) is independent from each other and can have different sizes. The detailed descriptions for both MultiDiscrete action spaces is shown in Tables \ref{tab:robots-act} and \ref{tab:factories-act}.

\begin{table}[h!]
\centering
\caption{Description of the robots' action space}
\label{tab:robots-act}
    \begin{tabular}{|l|l|c|c|}
        \hline
        \textbf{Component} & \textbf{Description} & \textbf{Range} \\
        \hline
        Source Unit & location of the selected unit to perform an action & [0, $h \times w - 1$] \\
        Action Type & NOOP, move, transfer, pickup, dig, self-destruct & [0, 5] \\
        Move param. & up, right, down, left & [0, 3] \\
        Transfer param. & center, up, right, down, left & [0, 4] \\
        Transfer amount & 25\%, 50\%, 75\%, 95\% & [0, 3] \\
        Transfer resource & ice, ore, power & [0, 2] \\
        \hline
    \end{tabular}
\end{table}

Table \ref{tab:robots-act} shows the action space for robts. In this way, an action is obtained by sampling a value from each of the components. However, note how the first component, the \textit{action type}, is the one that determines which action parameters are going to be used. For example, the action \textit{move - right - left - 25\% - ore} will execute \textit{move - right} and the action \textit{transfer - left - up - 50\% - ice} will execute \textit{transfer - up - 50\% - ice}. The actions of \textit{pickup}, \textit{dig} and \textit{self-destruct} don't need any other parameters as robots can only pick up a fixed amount of power, dig just means that the robot will dig whatever resource is in the tile where it's digging and self-destruct is self-explanatory.

\begin{table}[h!]
\centering
\caption{Description of the factories' action space}
\label{tab:factories-act}
    \begin{tabular}{|l|l|c|c|}
        \hline
        \textbf{Component} & \textbf{Description} & \textbf{Range} \\
        \hline
        Source Unit & location of the selected unit to perform an action & [0, $h \times w - 1$] \\
        Action Type & NOOP, build\_light, build\_heavy, grow\_lichen & [0, 3] \\
        \hline
    \end{tabular}
\end{table}

\section{Reward distributions} \label{app-reward}
RL agents must master several skills to have a chance to win in Lux, the most challenging one being efficiently generating water: each factory starts with 150 water and metal, and consumes 1 water at each turn. A factory is lost if it runs out of water, and can only generate water by sending robots to dig ice and transfer it to the factories to have the ice refined. If the RL agents don't master this sequence of actions, which requires a lot of efficient exploration, the games during training will last for 150 turns, when both players will lose all their factories. Solving this challenge has been the main stopper in our work, and for this reason, our initial dense reward only rewards robots for digging ice and factories for refining it. Ideally, after the RL agents master this strategy, they need to learn late-game dynamics, in which they have to spend water to grow lichen without running out of it. For this reason, we propose changing the reward distribution after lots of training updates to a sparse reward indicating the lichen amount at the last turn of the game. The latter can create a curriculum for RL agents to master basic skills and transfer them to late-game dynamics.

\section{Self-play} \label{app-selfplay}
Self-play enables shaping the learning dynamics in favour of the learner agents. There is a whole subfield of MARL exploring different strategies for training populations of diverse agents that are helpful to train state-of-the-art agents to play complex video games. Population-based training \citep{garnelo2021pick} is about defining a metric to rank the true skill of agents within a population and a strategy to sample opponents for each agent in the pool at each game. We argue that population-based training has been a key feature of the recent most impactful works in MARL \citep{vinyals2019grandmaster, berner2019dota}. However, it is usually overlooked and is sometimes obscure, because there is a general lack of frameworks and sources to implement such techniques. 

In this work, we have also run some experiments using the TrueSkill \citep{graepel2007bayesian} metric to rank the pool of agent's checkpoints. Instead of using the pool as a queue and keep removing the oldest checkpoints, using such a metric allows us to fairly evaluate whether the old checkpoints that were to be removed were actually less skilled than the others. However, we haven't used such a metric in our sweeping experiments because we haven't been able to scale the ranking function across the parallel jobs used due to our dependency on the external library \footnote{\url{https://trueskill.org/}}.

\section{Experiments} \label{app-performance}

\begin{figure}[h!]
\includegraphics[width=\linewidth]{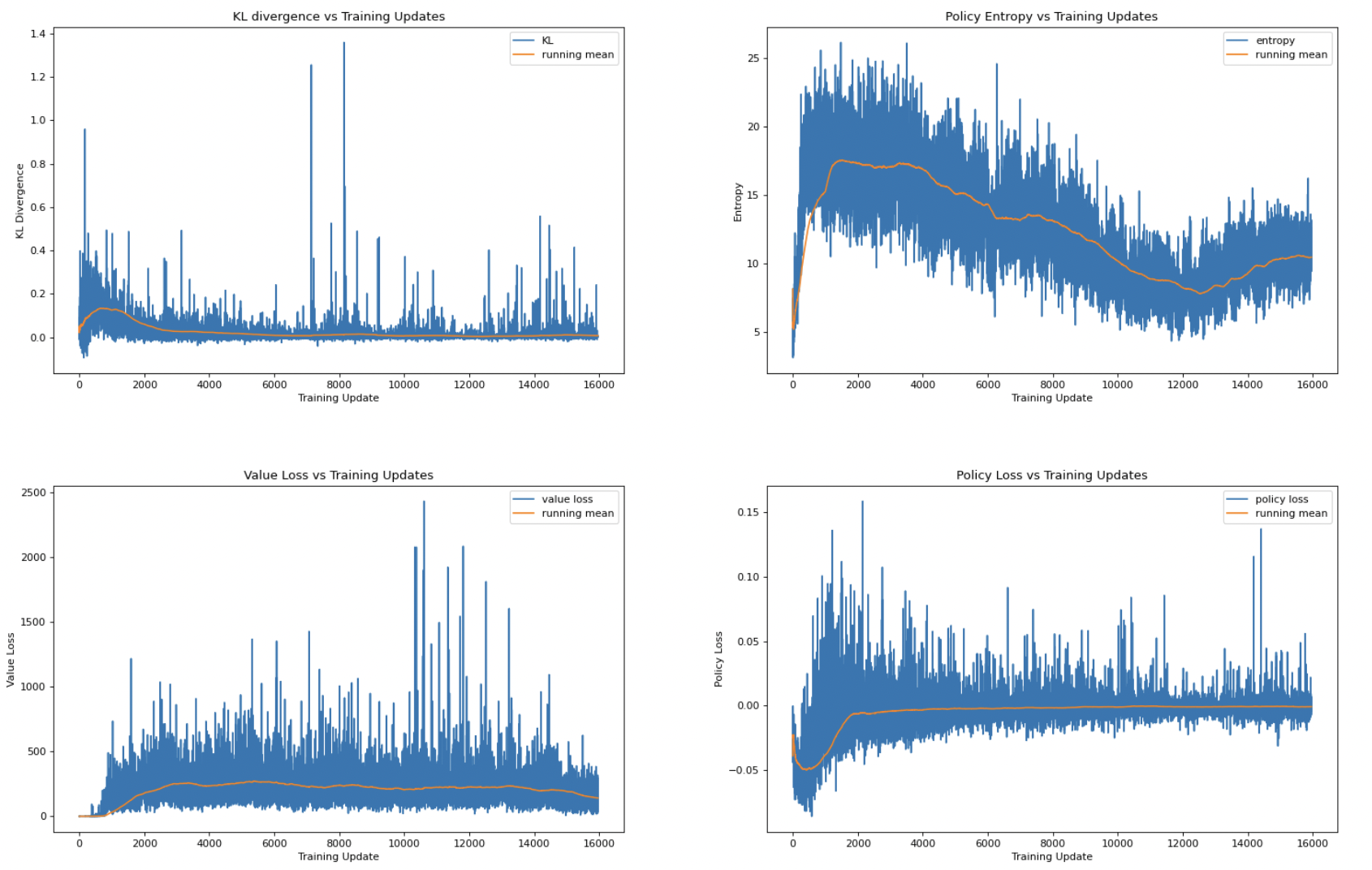}
\centering
\label{fig:performance}
\end{figure}

Figure \ref{fig:performance} shows the complete logs of the final PPO run. The architecture in Figure \ref{arch} was trained with PPO and run for 220M environment steps, which took 126 hours using 8 Nvidia A600 GPUs. Figure \ref{performance} indicates:

\begin{itemize}
    \item \textbf{The value loss increases during training}. The spikes in the value loss are usually a good sign since it means that the agent observes transitions in which it is unable to estimate the value correctly at first. The latter indicates that the agent is exploring (possibly) high-value behaviours. However, at convergence, we would expect the value loss to get closer to 0.
    \item \textbf{The entropy increases at the beginning and decreases during training}. That is usually a good sign and means that the policy converges to a less stochastic behaviour. 
    \item \textbf{The policy loss gets closer to 0 during training}. That is usually a good sign and is correlated with the decrease in entropy. It means that the policy changes less abruptly across updates, hence converging to a less stochastic policy.
    \item \textbf{The KL divergence gets close to 0 but spikes abruptly}. The abrupt spikes are often undesired and indicate that the policy takes big learning steps, which can cause forgetting previously learnt behaviours. Ideally, we would like the KL to remain similar during training and close to 0.
\end{itemize}

\end{appendices}

\end{document}